\documentclass[runningheads]{llncs}
\usepackage{graphicx}
\usepackage{subfig}
\usepackage{xcolor}
\usepackage{mathtools}
\usepackage{bm}
\usepackage{amsmath}
\usepackage{amssymb}
\usepackage{aliascnt}
\newcommand{\mean}[1]{\ensuremath{\operatorname{ave}(#1)}}

\usepackage{cite}
\usepackage{hyperref}
\usepackage{url}

\makeatletter
\AtBeginDocument{%
  \def\doi#1{\url{https://doi.org/#1}}}
\makeatother

\begin{document}
\title{Per-run Algorithm Selection with Warm-starting using Trajectory-based Features}
\titlerunning{Per-run Algorithm Selection with Warm-starting}

\author{Ana Kostovska\inst{1,2},
Anja Jankovic\inst{3},
Diederick Vermetten\inst{4}, 
Jacob de Nobel\inst{4},
Hao Wang\inst{4}, 
Tome Eftimov\inst{1}, 
Carola Doerr\inst{3}}

\authorrunning{A. Kostovska, A. Jankovic, D. Vermetten, et al.}

\institute{
Jo\v{z}ef Stefan Institute, Ljubljana, Slovenia
\and 
Jo\v{z}ef Stefan International Postgraduate School, Ljubljana, Slovenia
\and 
LIP6, Sorbonne Universit\'e, CNRS, Paris, France
\and 
LIACS, Leiden University, Leiden, The Netherlands
}

\maketitle  

\begin{abstract}
Per-instance algorithm selection seeks to recommend, for a given problem instance and a given performance criterion, one or several suitable algorithms that are expected to perform well for the particular setting. The selection is classically done offline, using openly available information about the problem instance or features that are extracted from the instance during a dedicated feature extraction step. This ignores valuable information that the algorithms accumulate during the optimization process.

In this work, we propose an alternative, online algorithm selection scheme which we coin as ``per-run'' algorithm selection. In our approach, we start the optimization with a default algorithm, and, after a certain number of iterations, extract instance features from the observed trajectory of this initial optimizer to determine whether to switch to another optimizer. We test this approach using the CMA-ES as the default solver, and a portfolio of six different optimizers as potential algorithms to switch to. In contrast to other recent work on online per-run algorithm selection, we warm-start the second optimizer using information accumulated during the first optimization phase. We show that our approach outperforms static per-instance algorithm selection. We also compare two different feature extraction principles, based on exploratory landscape analysis and time series analysis of the internal state variables of the CMA-ES, respectively. We show that a combination of both feature sets provides the most accurate recommendations for our test cases, taken from the BBOB function suite from the COCO platform and the YABBOB suite from the Nevergrad platform.
\end{abstract}

\keywords{Algorithm Selection \and
Black-Box Optimization  \and
Exploratory Landscape Analysis \and Evolutionary Computation}

\section{Introduction}
\label{sec:intro}

\begin{figure}[t]
    \centering
    \includegraphics[width=0.95\textwidth]{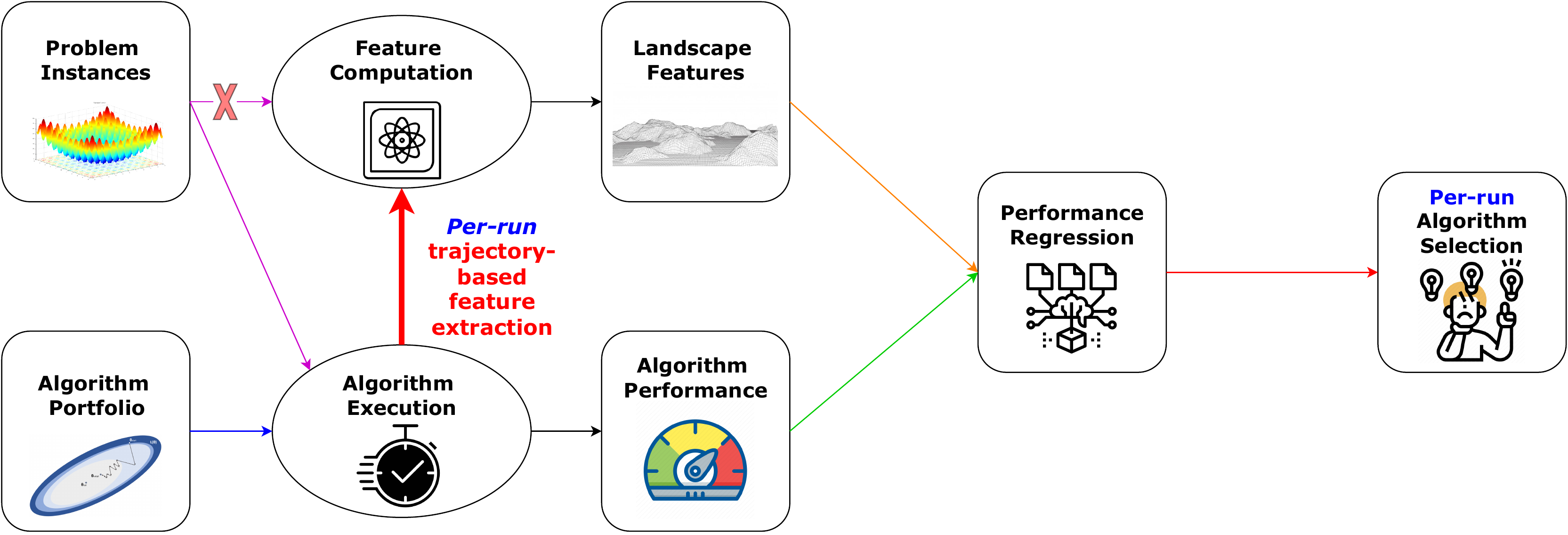}
    \caption{Per-run algorithm selection pipeline. The overhead cost of computing ELA features per problem instance is circumvented via collecting information about the instance during the default optimization algorithm run.}
    \label{fig:per-run-as-full}
\end{figure}

It is widely known that optimization problems are present in many areas of science and technology. A particular subset of these problems are the black-box problems, for which a wide range of optimization algorithms has been developed. However, it is not always clear which algorithm is the most suitable one for a particular problem. Selecting which algorithm to use comes with its own cost and challenges, so the choice of an appropriate algorithm poses a meta-optimization problem that has itself become an increasingly important area of study.\\
Moreover, a user needs to be able to select different algorithms for different \emph{instances} of the same problem, which is a scenario that very well reflects real-world conditions. This \textbf{per-instance algorithm selection} most often relies on being able to compute a set of features which capture the relevant properties of the problem instance at hand. A popular approach is the \emph{landscape-aware} algorithm selection, where the problem features' definition stems from the field of \emph{exploratory landscape analysis} (ELA)~\cite{mersmann_exploratory_2011}. In this approach, an initial set of points is sampled and evaluated on the problem instance to identify its global properties. However, this induces a significant overhead cost to the algorithm selection procedure, since the initial sample of points used to extract knowledge from the problem instance is usually not directly used by the chosen algorithm in the subsequent optimization process.\\
Previous research into landscape-aware algorithm selection suggests that, as opposed to creating a separate set of samples to compute ELA features in a dedicated preprocessing step, one could use the samples observed by some initial optimization algorithm. This way, the algorithm selection changes from being a purely offline procedure into being one which considers whether or not to switch between different algorithms during the search procedure. This is an important step towards dynamic (online) algorithm selection, in which the selector is able to track and adapt the choice of the algorithm throughout the optimization process in an intelligent manner.\\
In this paper, we coin the term \textbf{per-run algorithm selection} to refer to the case where we make use of information gained by running an initial optimization algorithm (\emph{A1}) during a single run to determine which algorithm should be selected for the remainder of the search. This second algorithm (\emph{A2}) can then be warm-started, i.e., initialized appropriately using the knowledge of the first one. The pipeline of the approach is shown in Fig.~\ref{fig:per-run-as-full}.\\
Following promising results from~\cite{DBLP:conf/evoW/JankovicED21}, in this work we apply our \emph{trajectory-based algorithm selection} approach to a broader set of algorithms and problems. To extract relevant information about the problem instances, we rely on ELA features computed using samples and evaluations observed by the initial algorithm's search trajectory, i.e., \emph{local} landscape features. Intuitively, we consider the problem instance as perceived from the algorithm's viewpoint.\\
In addition, we make use of an alternative aspect that seems to capture critical information during the search procedure -- the algorithm's internal state, quantified through a set of state variables at every iteration of the initial algorithm. To this end, we choose to track their evolution during the search by computing their corresponding \emph{time-series} features.\\
Using the aforementioned values to characterize problem instances, we build algorithm selection models based on the prediction of the fixed-budget performance of the second solver on those instances, for different budgets of function evaluations. We train and test our algorithm selectors on the well-known BBOB problem collection of the COCO platform~\cite{cocoplat}, and extend the testing on the YABBOB collection of the Nevergrad platform~\cite{nevergrad}. We show that our approach leads to promising results with respect to the selection accuracy and we also point out interesting observations about the particularities of the approach.\\
\textbf{State of the Art.} Given an optimization problem, a specific instance of that problem which needs to be solved, and a set of algorithms which can be used to solve it, the so-called per-instance algorithm selection problem arises. How does one determine which of those algorithms can be expected to perform best on that particular instance? In other words, one is not interested in having an algorithm recommendation for a whole problem class (such as TSP or SAT in the discrete domain), but for a specific instance of some problem. A large body of work exists in this line of research~\cite{BischlMTP12, CossonDLATZ21, HutterKV19, KerschkeHNT19, LindauerHHS15, XuHHL12}. All of these deal predominantly with offline AS. An effort towards online AS has been recently proposed~\cite{meidani2022online}, where the switching rules between algorithms were defined based on non-convex ratio features extracted during the optimization process. However, this particular study is not based on using supervised machine learning techniques to define the switching rule, which is the key difference presented in our approach.\\
\textbf{Paper Outline.} In Section~\ref{sec:data-coll}, we introduce the problem collections and the algorithm portfolio, and give details about the raw data generation for our experiments. The full experimental pipeline is more closely presented in Section~\ref{sec:exp-setup}. We discuss the main results on two benchmark collections in Sections~\ref{sec:res-coco} and~\ref{sec:res-ng}, respectively. Finally, Section~\ref{sec:conclusions} gives several possible avenues for future work.\\
\textbf{Data and Code Availability.} Our source code, raw data, intermediate artefacts and analysis scripts have been made available on our Zenodo repository~\cite{zenodo_ppsn_algoselect}. In this paper, we highlight only selected results for reasons of space.

\section{Data Collection}
\label{sec:data-coll}

\begin{figure}
    \centering
    \includegraphics[width=0.97\textwidth,trim=0mm 5mm 0mm 5mm,clip]{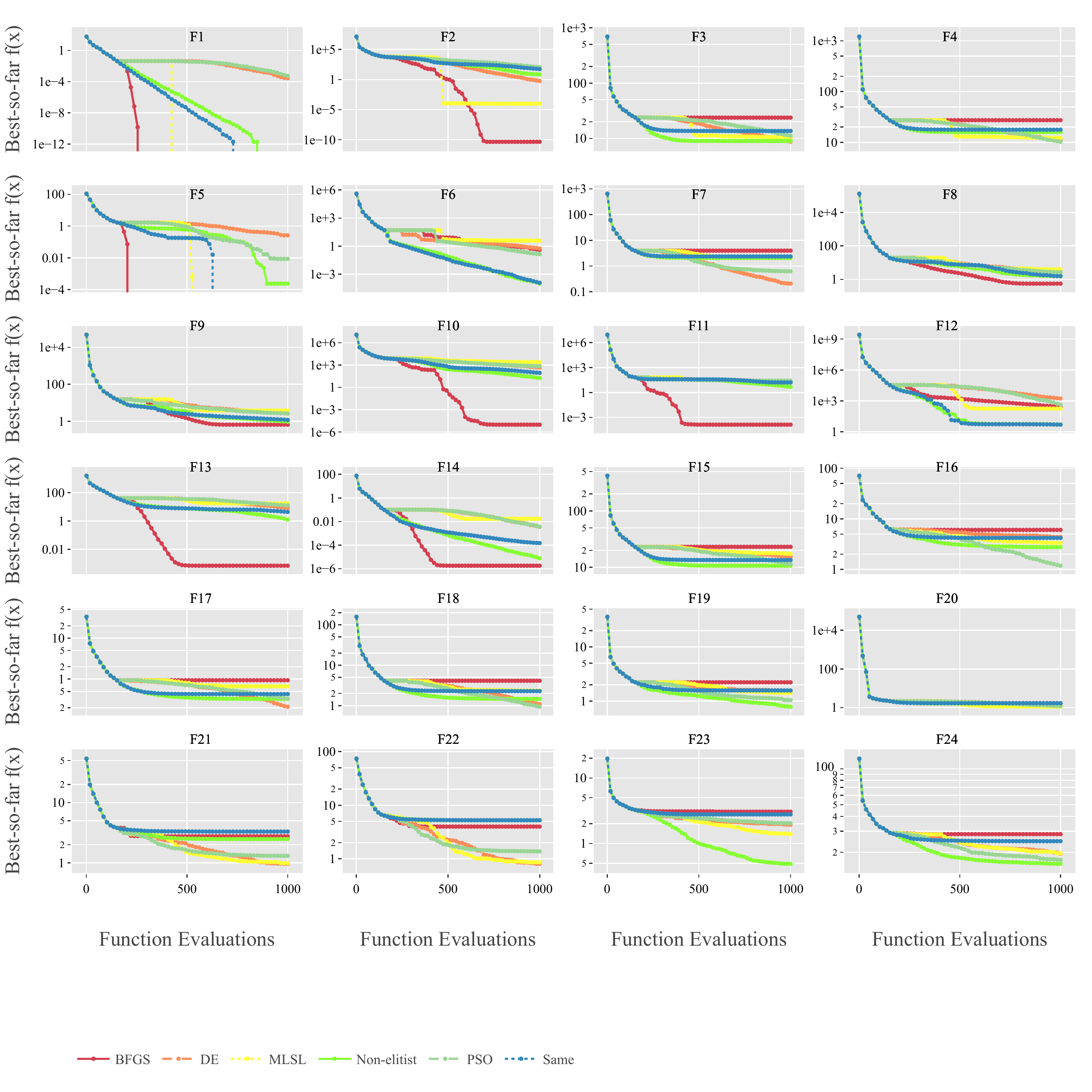}
    \caption{Mean best-so-far function value (precision to global optimum) for each of the six algorithms in the portfolio. For computational reasons, each line is calculated based on a subset of 10 runs on each of the 10 instances used, for a total of 100 runs. Note that the first 150 evaluations for each algorithm are identical, since this is the budget used for A1. Figure generated using IOHanalyzer~\cite{IOHanalyzer}.}
    \label{fig:perf_matrix_5D_30B}
\end{figure}

\begin{figure}
    \centering
    \includegraphics[width=\textwidth,trim=10mm 115mm 10mm 125mm,clip]{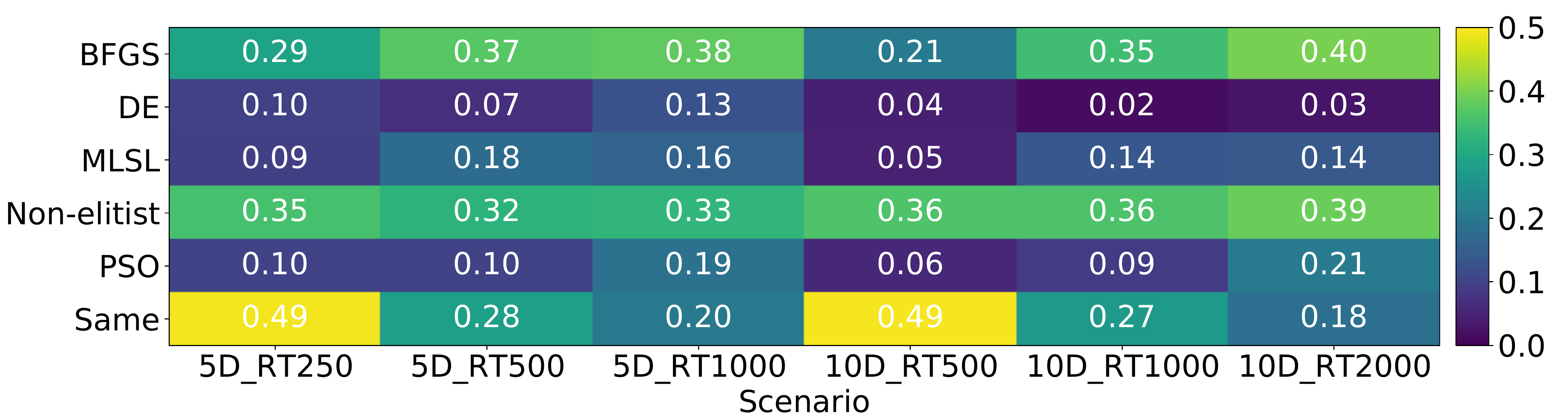}
    \caption{Matrix showing for each scenario (with respect to the dimensionality and A2 budget) in what proportion of runs each algorithm reaches the best function value. Note that these value per scenario can add to more than 1 because of ties.}
    \label{fig:best_alg}
\end{figure}

\textbf{Problem Instance Portfolio.} To implement and verify our proposed approach, we make use of a set of black-box, single-objective, noiseless problems. The data set is the BBOB suite from the COCO platform~\cite{cocoplat}, which is a very common benchmark set within numerical optimization community. This suite consists of a total of 24 functions, and each of these functions can be changed by applying pre-defined transformations to both its domain and objective space, resulting in a set of different instances of each of these problems that share the same global characteristics~\cite{bbob-functions}.\\
Another considered benchmark set is the YABBOB suite from the Nevergrad platform~\cite{nevergrad}, that contains 21 black-box functions, out of which we keep 17 for this paper. By definition, YABBOB problems do not allow for generating different instances.\\ 
\textbf{Algorithm Portfolio.}
As our algorithm portfolio, we consider the one used in~\cite{arxiv_algoselect_cec, arxiv_warmstart_schroder}. This gives us a set of 5 black-box optimization algorithms: MLSL~\cite{mlsl1, mlsl2}, BFGS~\cite{bfgs1, bfgs2, bfgs3, bfgs4}, PSO~\cite{PSO}, DE~\cite{de} and CMA-ES~\cite{cmaes}. Since for the CMA-ES we consider two versions from the modular CMA-ES framework~\cite{modCMA} (elitist and non-elitist), this gives us a total portfolio of 6 algorithm variants. The implementation of the algorithms used can be found in more detail in our repository~\cite{zenodo_ppsn_algoselect}.
\textbf{Warm-starting.} 
To ensure we can switch from our initial algorithm (A1) to any of the others (A2), we make use of a basic warm-starting approach specific to each algorithm. For the two versions of modular CMA-ES, we do not need to explicitly warm-start, since we can just continue the run with the same internal parameters and turn on elitist selection if required. The detailed warm-start mechanisms are discussed in~\cite{arxiv_algoselect_cec}, and the implementations are available in our repository~\cite{zenodo_ppsn_algoselect}.\\
\textbf{Performance Data.} 
For our experiments, we consider a number of data collection settings, based on the combinations of dimensionality of the problem, where we use both 5- and 10-dimensional versions of the benchmark functions, and budget for A1, where we use $30\cdot D$ budget for the initial algorithm. This is then repeated for all functions of both the BBOB and the YABBOB suite. For BBOB, we collect 100 runs on each of the first 10 instances, resulting in $1\,000$ runs per function. For YABBOB (only used for testing), we collect 50 runs on each function (due to no instances in Nevergrad). \\
In Figure~\ref{fig:perf_matrix_5D_30B}, we show the performance of the six algorithms in our portfolio in the 5-dimensional case. Since the A1 budget is $30\cdot D=150$, the initial part of the search is the same for all algorithms until this point. In the figure, we can see that, for some functions, clear differences in performance between the algorithm appear very quickly, while for other functions the difference only becomes apparent after some more evaluations are used. This difference leads us to perform our experiments with three budgets for the A2 algorithm, namely $20\cdot D$, $70\cdot D$ and $170\cdot D$.\\
To highlight the differences between the algorithms for each of these scenarios, we can show in what fraction of runs each algorithm performs best. This is visualized in Figure~\ref{fig:best_alg}. Here we can see that while some algorithms are clearly more impactful than others, the differences between them are still significant. This indicates that there would be a significant difference between a virtual best solver which selects the best algorithm for each run and a single best solver which uses only one algorithm for every run.

\section{Experimental Setup}
\label{sec:exp-setup}

\textbf{Adaptive Exploratory Landscape Analysis.} 
As previously discussed, the \emph{per-run} trajectory-based algorithm selection method consists of extracting ELA features from the search trajectory samples during a single run of the initial solver. A vector of numerical ELA feature values is assigned to each run on the problem instance, and can be then used to train a predictive model that maps it to different algorithms' performances on the said run. To this end, we use the ELA computation library named \textsc{flacco}~\cite{flacco_CEC}.\\
Among over 300 different features (grouped in feature sets) available in \textsc{flacco}, we only consider features that do not require additional function evaluations for their computation, also referred to as \emph{cheap features}~\cite{BelkhirDSS17}. They are computed using the fixed initial sample, while \emph{expensive features}, in contrast, need additional sampling during the run, an overhead that makes them more inaccessible for practical use. For the purpose of this work, as suggested in preliminary studies~\cite{DBLP:conf/evoW/JankovicED21, arxiv_algoselect_cec}, we use 38 cheap features most commonly used in the literature, namely those from \emph{y-Distribution}, \emph{Levelset}, \emph{Meta-Model}, \emph{Dispersion}, \emph{Information Content} and \emph{Nearest-Better Clustering} feature sets.\\
We perform this per-run feature extraction using the initial $A1 = 30 \cdot D$ budget of samples and their evaluations per each run of each of the first 10 instances of each of the 24 BBOB problems, as well as 17 YABBOB problems (that have no instances) in dimensions 5 and 10.\\
\textbf{Time-Series Features.} 
In addition to ELA features computed during the optimization process, we consider an alternative -- \emph{time-series} features of the internal states of the CMA-ES algorithm. Since the internal variables of an algorithm are adapted during the optimization, they could potentially contain useful information about the current state of the optimization. Specifically, we consider the following internal variables: the step size $\sigma$, the eigenvalues of covariance matrix $\vec{v}$, the evolution path $\vec{p}_c$and its conjugate $\vec{p}_\sigma$, the Mahalanobis distances from each search point to the center of the sampling distribution $\vec{\gamma}$, and the log-likelihood of the sampling model $\mathcal{L}\left(\vec{m}, \sigma^2, \mathbf{C}\right)$. We consider these dynamic strategy parameters of the CMA-ES as a multivariate real-valued time series, for which at every iteration of the algorithm, we compute one data point of the time series as follows: $\forall t\in[L]$:\\
 $\vec{\psi}_t :=\left(\sigma, \mathcal{L}(\vec{m}, \sigma^2, \mathbf{C}), ||\vec{v}||, ||\vec{p}_\sigma||, ||\vec{p}_c||, ||\vec{\gamma}||, \mean{\vec{v}},\mean{\vec{p}_\sigma},\mean{\vec{p}_c},\mean{\vec{\gamma}}\right)^\top, $\\
where $L$ represents the number of iterations these data points were sampled, which equals the A1 budget divided by the population size of the CMA-ES. In order to store information invariant to the problem dimension, we compute the component-wise average $\mean{\cdot}$ and norm $||\vec{x}|| = \sqrt{\vec{x}^\top\vec{v}}$ of each vector variable.\\
Given a set of $m$ feature functions $\left\{\phi_i\right\}_{i=1}^m$ from \textsc{tsfresh} (where $\phi_i \colon \mathbb{R}^L \rightarrow \mathbb{R}$), we apply each feature function over each variable in the collected time series. Examples of such feature functions are autocorrelation, energy and continuous wavelet transform coefficients. In this paper, we take this entire time series (of length $L$) as the feature window. We employ all $74$ feature functions from the \textsc{tsfresh} library~\cite{tsfresh}, to compute a total of $9\,444$ time-series features per run. After the feature generation, we perform a feature selection method using a Random Forests classifier trained to predict the function ID, for computing the feature importance. We then select only the features whose importance is larger than $2\times 10^{-3}$. This selection procedure yields $129$ features, among which features computed on the Mahalanobis distance and the step-size $\sigma$ are dominant. More details on this approach can be found in~\cite{timeseriespaper}.\\
\textbf{Regression Models.} 
To predict the algorithm performance after the A2 budget, we use as performance metric the target precision reached by the algorithm in the fixed-budget context (i.e., after some fixed number of function evaluations). We create a mapping between the input feature data, which can be one of the following: (1)
    the trajectory-based representation with 38 ELA features per run (ELA-based AS), 
    (2) the trajectory-based representation with 129 time-series (TS) features per run (TS-based AS), or
    (3) a combination of both (ELA+TS-based AS), 
and the target precision of different algorithm runs. We then train supervised machine learning (ML) regression models that are able to predict target precision for different algorithms on each of the trajectories involved in the training data. Following some strong insights from~\cite{DBLP:conf/gecco/JankovicD20} and subsequent studies, we aim at predicting the logarithm ($log10$) of the target precision, in order to capture the order of magnitude of the distance from the optimum. In our case, since we are dealing with an algorithm portfolio, we have trained a separate single target regression (STR) model for each algorithm involved in our portfolio. We opt for using a random forest (RF) regression, as previous studies have shown that it provides promising results for automated algorithm performance prediction~\cite{DBLP:conf/gecco/JankovicPED21}. To this end, we use the RF implementation from the Python package \textsc{scikit-learn}~\cite{pedregosa2011scikit}.\\
\textbf{Evaluation Scenarios.} 
To find the best RF hyperparameters and to evaluate the performance of the algorithm selectors, we have investigated two evaluation scenarios:\\
    \textbf{(1) Leave-instance out validation:} in this scenario, 70\% of the instances from each of the 24 BBOB problems are randomly selected for training and 30\% are selected for testing. Put differently, all 100 runs for the selected instance will either appear in the training or the test set. We thus end up with 16\,800 trajectories used for training and 7\,200 trajectories for testing.\\
    \textbf{(2) Leave-run out validation}: in this scenario, 70\% of the runs from each BBOB problem instance are randomly selected for training and 30\% are selected for testing. Again, we end up with 16\,800 trajectories used for training and 7\,200 trajectories for testing.\\
We repeat each evaluation scenario five independent times, in order to analyze the robustness of the results. Each time, the training data set was used to find the best RF hyperparameters, while the test set was used only for evaluation of the algorithm selector. \\
\textbf{Hyperparameter Tuning for the Regression Models.} 
The best hyperparameters are selected for each RF model via grid search for a combination of an algorithm and a fixed A2 budget. The training set for finding the best RF hyperparameters for each combination of algorithm and budget is the same. Four different RF hyperparameters are selected for tuning: (1) \emph{n\_estimators}: the number of trees in the random forest; (2) \emph{max\_features}: the number of features used for making the best split; (3) \emph{max\_depth}: the maximum depth of the trees, and (4) \emph{min\_samples\_split}: the minimum number of samples required for splitting an internal node in the tree. The search spaces of the hyperparameters for each RF model utilized in our study are presented in Table~\ref{tab:hyperparameters}.
\begin{table}[t]
\begin{center}
\caption{RF hyperparameter names and their corresponding values considered in the grid search.}
\label{tab:hyperparameters}
\begin{tabular}{ cc }
\hline
 Hyperparameter & Search space \\ 
\hline
 n\_estimators & $[100, 300]$\\
 max\_features & $[\textsc{auto}, \textsc{sqrt}, \textsc{log2}]$ \\ 
 max\_depth & $[3,5,15, \textsc{None}]$ \\
 min\_samples\_split & $[2, 5, 10]$ \\
 \hline
\end{tabular}
\end{center}
\end{table}\\
\textbf{Per-run Algorithm Selection.} 
In real-world dynamic AS applications, we rely on the information obtained within the current run of the initial solver on a particular problem instance to make our decision to switch to a better suited algorithm. A randomized component of black-box algorithms comes into play here, as one algorithm's performance can vastly differ from one run to another on the very same problem instance. \\
We estimate the quality of our algorithm selectors by comparing them to standard baselines, the virtual best solver (VBS) and the single best solver (SBS). As we make a clear distinction between per-run and per-instance perspective, in order to compare we need to suitably aggregate the results. Our baseline is the \emph{per-run VBS}, which is the selector that always chooses the real best algorithm for a particular run on a certain problem (i.e., function) instance. We then define $VBS_{iid}$ and $VBS_{fid}$ as virtual best solvers on instance and problem levels, i.e., selectors that always pick the real best algorithm for a certain instance (across all runs) or a certain problem (across all instances). Last, we define the SBS as the algorithm that is most often the best one across all runs.\\
For each of these methods, we can define their performance relative to the \emph{per-run VBS} by considering their performance ratio, which is defined on each run as taking the function value achieved by the VBS and dividing it by the value reach by the considered selector. As such, the performance ratio for the \emph{per-run VBS} is 1 by definition, and in $[0,1]$ for each other algorithm selector. \\
To measure the performance ratio for the algorithm selectors themselves, we calculate this performance ratio on every run in the test-set of each of the 5 folds, and average these values. We point out here that the performance of different AS models are not statistically compared, since the obtained performance values from the folds are not independent~\cite{demvsar2006statistical}.

\section{Evaluation Results: COCO}
\label{sec:res-coco}

\begin{figure}[t]
    \centering
    \includegraphics[width=\textwidth,trim=0mm 62mm 0mm 80mm,clip]{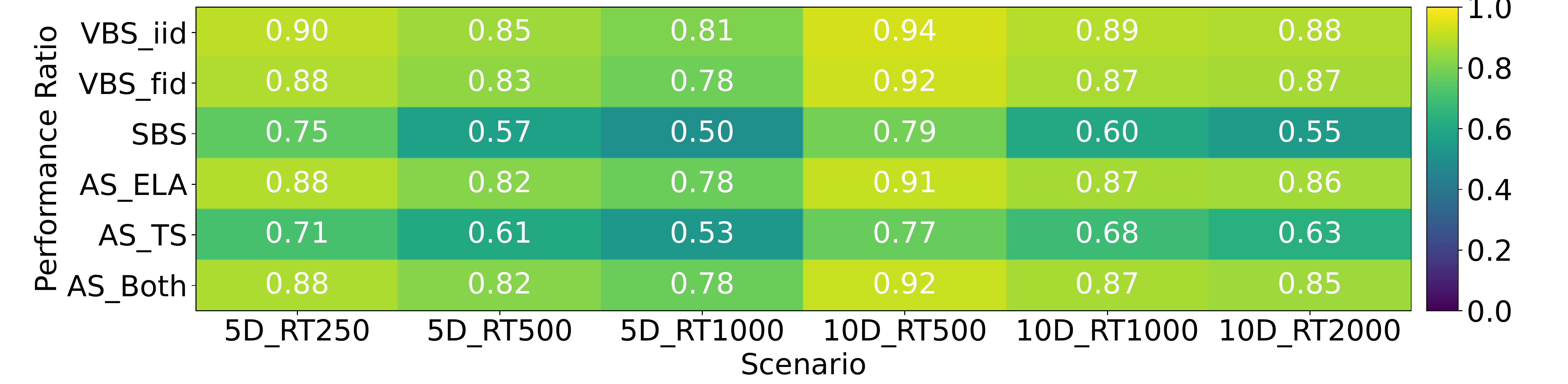}
    \caption{Heatmap showing for each scenario the average performance ratio relative to the per-run virtual best solver of different versions of VBS, SBS and algorithm selectors (based on the per-instance folds). Scenario names show the problem dimensionality and the total used budget.}
    \label{fig:vbs_comp_with_as}
\end{figure}

\begin{figure}[t]
    \centering
    \includegraphics[width=\textwidth,trim=0mm 100mm 8mm 110mm,clip]{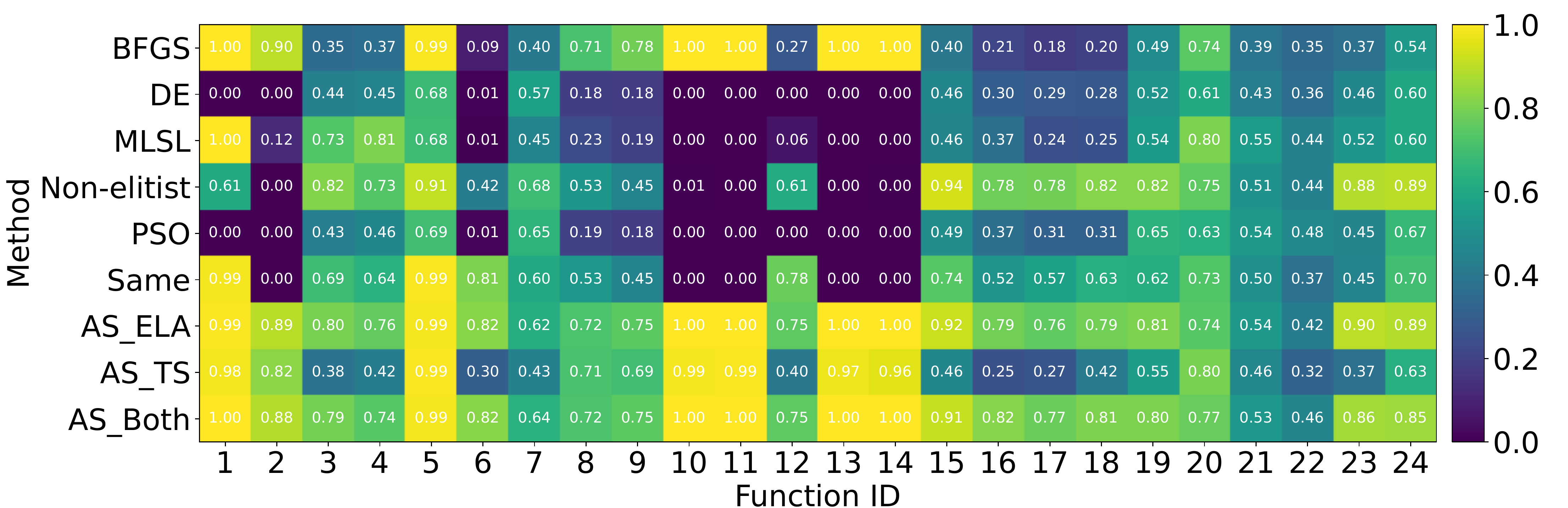}
    \caption{Heatmap showing for each 5-dimensional BBOB function the mean performance ratio at 500 total evaluations relative to the per-run virtual best solver, as well as the average performance ratio of each of the 3 algorithm selectors.}
    \label{fig:perf_ratios}
\end{figure}

For our first set of experiments, we train our algorithm selectors on BBOB functions using the evaluation method described in Section~\ref{sec:exp-setup}. Since we consider 2 dimensionalities of problems and 3 different A2 budgets, we have a total of 6 scenarios for each of the 3 algorithm selectors (ELA-, TS-, and ELA+TS-based). In Figure~\ref{fig:vbs_comp_with_as}, we show the performance ratios of these selectors, as well as the performance ratios of the previously described VBS and SBS baselines. Note that for this figure, we make use of the \emph{per-instance} folds, but results are almost identical for the \emph{per-run} case.\\
Based on Figure~\ref{fig:vbs_comp_with_as}, we can see that the ELA-based algorithm selector performs almost as well as the per-function VBS, which itself shows only minor performance differences to the per-instance VBS. We also notice that as the total evaluation budget increases, the performance of every selector deteriorates. This seems to indicate that as the total budget becomes larger, there are more cases where runs on the same instance have different optimal switches. \\
To study the performance of the algorithm selectors in more detail, we can consider the performance ratios for each function separately, as is visualized in Figure~\ref{fig:perf_ratios}. From this figure, we can see that for the functions where there is a clearly optimal A2, all algorithm selectors are able to achieve near-optimal performance. However, for the cases where the optimal A2 is more variable, the discrepancy between the ELA and TS-based algorithm selectors increases.

\section{Evaluation Results: Nevergrad}
\label{sec:res-ng}

\begin{figure}[t]%
    \centering
    \subfloat[\centering $5D$]{{\includegraphics[width=5cm]{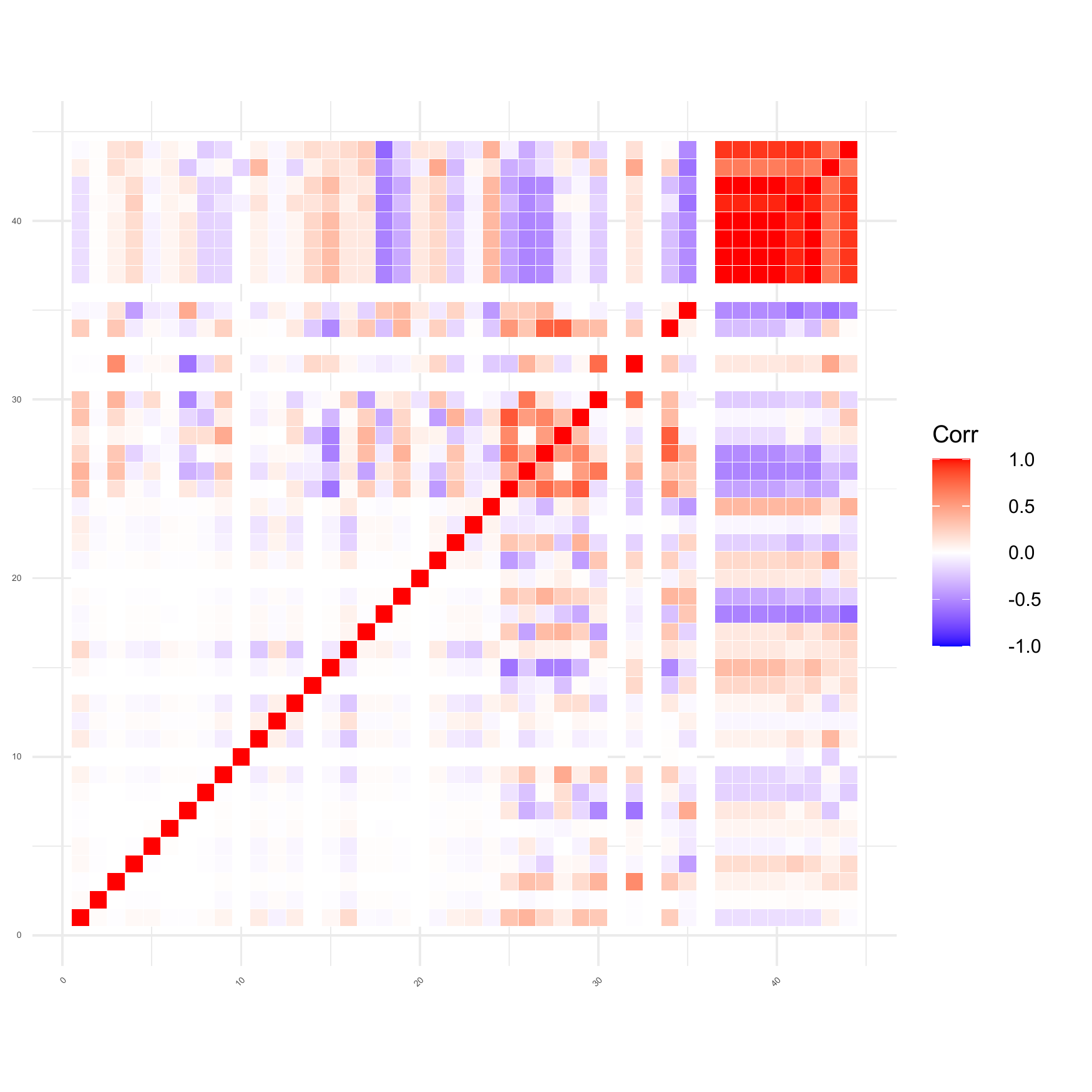} }}%
    \qquad
    \subfloat[\centering $10D$]{{\includegraphics[width=5cm]{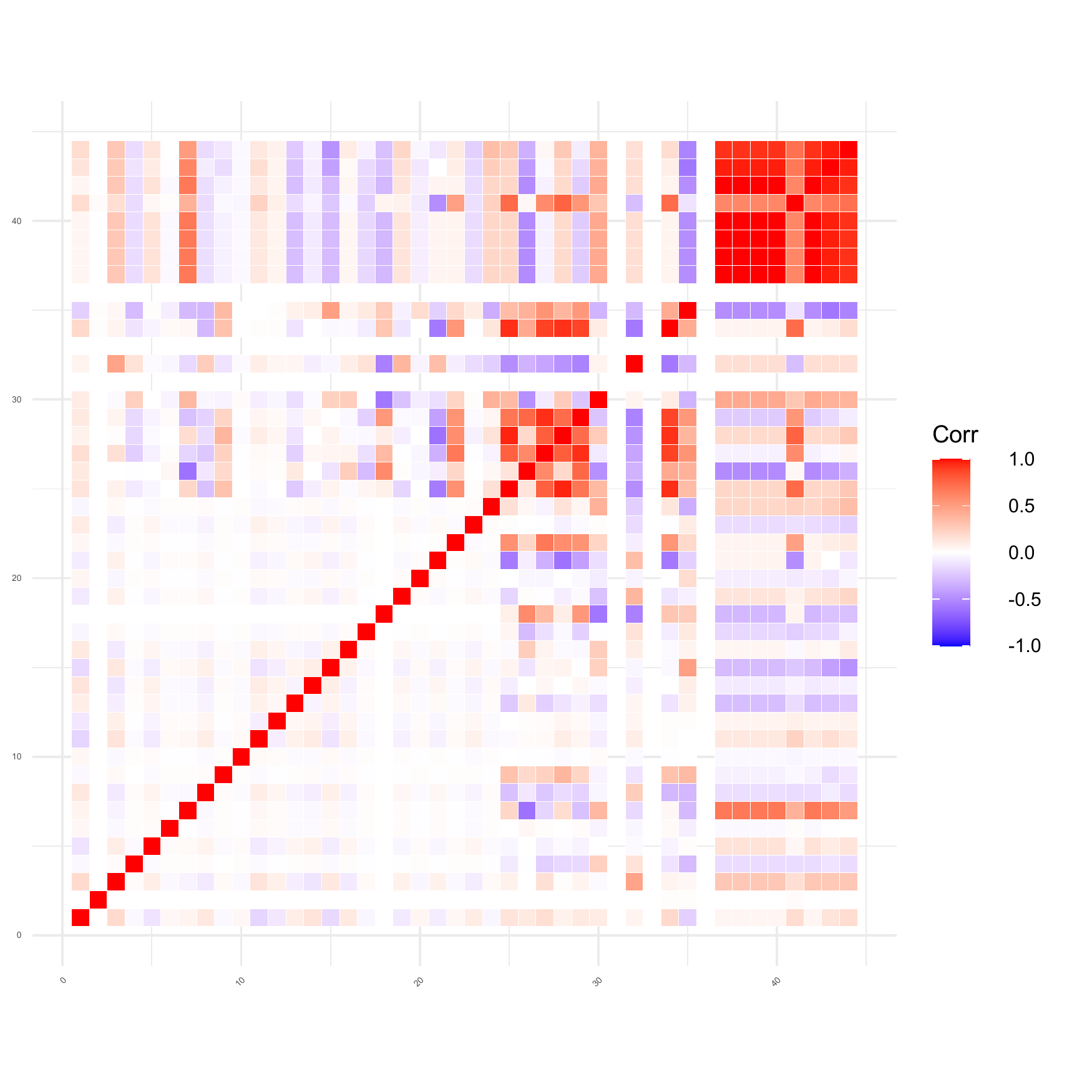} }}%
    \caption{Pearson correlation between BBOB (lower left portion, mostly white) and YABBOB trajectories for $5D$ and $10D$.}%
    \label{fig:complementarity}%
\end{figure}

We now study how a model trained on BBOB problem trajectories can be used to predict the performances on trajectories not included in the training. We do so by considering the YABBOB suite from the Nevergrad platform. While there is some overlap between these two problem collections, introducing another sufficiently different validation/test suite allows us to verify the stability of our algorithm selection models. 
We recall that for the performance data of the same algorithm portfolio on YABBOB functions, we have target precisions for $850$ runs, $50$ runs per 17 problems, in all considered A2 budgets.\\
\textbf{Training on COCO, testing on Nevergrad. }
This experiment has resulted in somewhat poorer performance of the algorithm selection models on an inherently different batch of problems. The comparison of the similarity between BBOB and YABBOB problems presented below 
nicely shows how the YABBOB problems are structurally more similar to one another than to the BBOB ones.
\begin{figure}[t!]
    \centering
    \includegraphics[width=\textwidth,trim=0mm 72mm 10mm 85mm,clip]{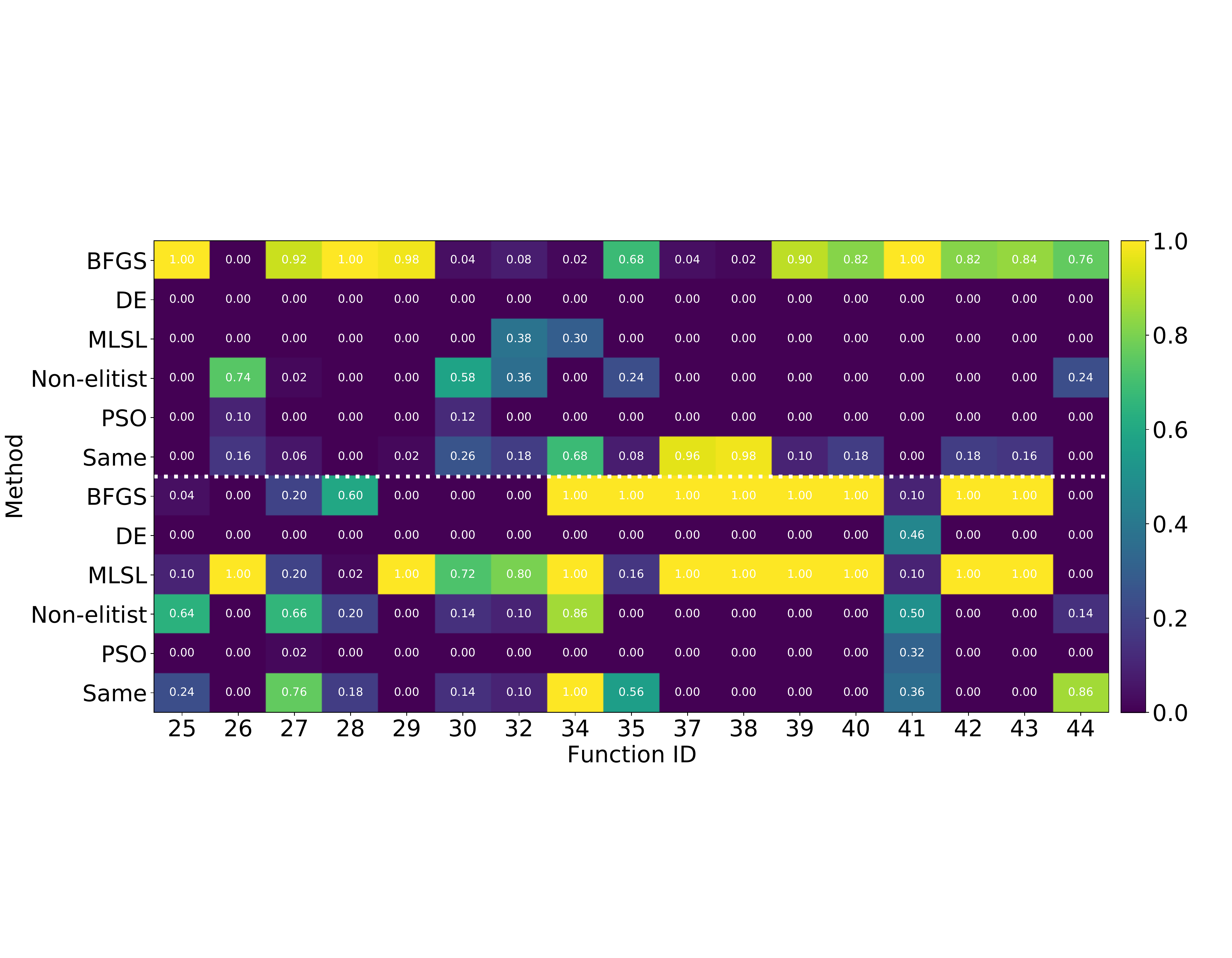}
    \caption{Heatmap showing for each 5-dimensional YABBOB/Nevergrad function the fraction of times each algorithm was optimal to switch to when considering a total budget of 500 evaluations (bottom) and how often each of these algorithm was selected by the algorithm selector trained on BBOB/COCO (top). Note that the columns of the bottom part can sum to more than 1 in case of ties.}
    \label{fig:ng_sel_alg_matrix}
\end{figure}
To investigate performance flaws of our approach when testing on Nevergrad, we compare, for each YABBOB problem, how often a particular algorithm is selected by the algorithm selection model trained on the BBOB data with how often that algorithm was actually the best one. This comparison is exhibited in Figure~\ref{fig:ng_sel_alg_matrix}. We observe that MLSL in particular is not selected often enough in the case of a large A2 budget, as well as a somewhat strong preference of the selector towards BFGS.
An explanation for these results may be the (dis)similarities between the benchmarks. Only for some YABBOB functions in the second half of the set we might have similarities in the trajectories already seen from the second half of the BBOB data, but this is anecdotal as the overall tendency is that there are few parallels between BBOB and YABBOB.\\
\textbf{Analyzing the complementarity between the COCO and Nevergrad suites. }
We illustrate the intra-similarity among the YABBOB test trajectories from the Nevergrad suite, which are not part of the training data set. This is shown via correlation between the YABBOB trajectories (test data) and the BBOB trajectories (training data). For this purpose, we first find the subspace that is covered by the training trajectories, where we then project the test trajectories. To find the subspace that is covered by training data, we apply singular value decomposition (SVD), following an approach presented in~\cite{eftimov2020linear}. For the training and test data, we summarize the trajectories on a problem level using the median values for each ELA feature by using all trajectory instances that belong to the same problem. Next, we map the BBOB trajectories to a linear vector space they cover (found by the SVD decomposition), where the trajectories are represented in different uncorrelated dimensions of the data. We then project each of the YABBOB trajectories to the linear subspace that is covered by the 24 BBOB problems, which allows us to find their correlation.\\
The Pearson correlation values between the trajectories obtained for $5D$ and $10D$ problem instances are showcased in Figure~\ref{fig:complementarity}. We opt for the Pearson correlation coefficient since the trajectories are projected in a linear subspace. The trajectories from 1 to 24 correspond to the BBOB suite, and the trajectories starting from 25 to 44 correspond to the YABBOB suite. It is important to recall here that the YABBOB problems F31, F33, F36 and F45 were omitted from further analysis due to missing values. This figure shows that the BBOB trajectories are not correlated (the white square portion of the lower left part of the heatmap), which confirms high diversity in the training trajectory portfolio. However, there are lower positive and negative correlations between BBOB and YABBOB trajectories, which indicate that the properties of the YABBOB trajectories are not captured in the training data. This might be a possible explanation for the poor performance for the algorithm selection models which is trained on the BBOB trajectories, but only tested on the YABBOB trajectories.

\section{Conclusions and Future Work}
\label{sec:conclusions}

We have shown the feasibility of building an algorithm selector based on a very limited amount of samples from an initial optimization algorithm. Results within the BBOB benchmark suite show performance comparable to the per-function virtual best solver when using a selector based on ELA features. While these results did not directly transfer to other benchmark suites, this seems largely caused by the relatively low similarity between the collections.\\ 
Since this work is based on warm-starting the algorithms using the information of the initial search trajectory, further improvement in warm-starting would be highly beneficial to the overall performance of this feature-based selection mechanism. In addition, identifying exactly what features contribute to the decisions being made can show us what properties might be important to the performance of the switching algorithm, which in turn can support the development of better warm-starting mechanisms.\\
While the time-series based approach did not perform as well as the one based on ELA, it still poses an interesting avenue for future research. In particular, it would be worthwhile to consider the combined model in more detail, and aim to identify the level of complementarity between landscape and algorithm state features, which would help gain insight into the complex interplay between problems and algorithms. 

\vspace{0.7ex}
\small{
\textbf{Acknowledgments.} 
The authors acknowledge financial support by the Slovenian Research Agency (research core grants No. P2-0103 and P2-0098, project grant No. N2-0239, and young researcher grant No. PR-09773 to AK), by the EC (grant No. 952215 - TAILOR), by the Paris Ile-de-France region (DIM RFSI project AlgoSelect), and by the CNRS INS2I institute (the RandSearch project). }

\bibliographystyle{splncs04}
\bibliography{references}

\end{document}